\documentclass[11pt]{article}

\usepackage[final]{acl}

\usepackage{times}
\usepackage{latexsym}
\usepackage[T1]{fontenc}
\usepackage[utf8]{inputenc}
\usepackage{microtype}
\usepackage{inconsolata}
\usepackage{graphicx}
\usepackage{amsmath}
\usepackage{amssymb}
\usepackage{booktabs}
\usepackage{multirow}
\usepackage{xcolor}
\usepackage{tikz} \usetikzlibrary{positioning, calc}
\usepackage{makecell}
\usepackage{float}
\usepackage{comment}
\usepackage{pifont}
\usepackage[most]{tcolorbox}
\usepackage{soul}

\usepackage{latexsym}
\usepackage{inconsolata}

\newcolumntype{C}[1]{>{\centering\arraybackslash}p{#1}}

\newcommand{\todo}[1]{}
\newcommand{\ks}[1]{}
\newcommand{\tor}[1]{}
\newcommand{\yp}[1]{}
\newcommand{\zyc}[1]{}
\newcommand{\rs}[1]{}

\title{DI-Bench: Systematically Generating In-Domain Data Intelligence Benchmarks for Enterprise Agents}

\author{
 \textbf{Jiangyun Zhang},
 \textbf{Kristen Surrao},
 \textbf{Torpong Nitayanont},
 \textbf{Yupei Zhang},
\\
 \textbf{Roopali Singh},
 \textbf{Zhiyu Chen},
 \textbf{Julia Huang},
 \textbf{Zhou Tang},
 \\
 \textbf{Shayan Ali Akbar},
 \textbf{Omar Alonso},
 \textbf{Erwin Cornejo},
 \textbf{Yuan Li},
 \textbf{Yi Zhang}
\\
 Amazon.com
\\
 \small{
   \textbf{Correspondence:} \href{mailto:email@domain}{zjiangyu@amazon.com}
 }
}

\begin{document}
\maketitle

\begin{abstract}

Evaluating enterprise agents on domain-specific benchmarks is critical, yet public benchmarks rarely evaluate whether agents can integrate business knowledge with analytical computation, and constructing such benchmarks manually is costly. We present DI-Bench, a pipeline for generating realistic benchmarks for data intelligence (DI), the practice of extracting insights from large volumes of enterprise data. To emulate realistic DI tasks that require both computation and knowledge retrieval, DI-Bench builds an \emph{artifact linkage graph} over data tables, dimensions, metrics, and documents to form questions involving structured data and associated knowledge. Ground truth answers are derived via query execution, followed by LLM question generation and validation. Applied to two public datasets, the pipeline produces a 731-task benchmark covering knowledge retrieval, analytical computation, and rule-grounded reasoning. To show the discriminatory capability and difficulty of the benchmark, we evaluate four models, revealing a substantial finding: models achieve only 32\% accuracy when doing computational tasks where retrieved business rules modify the computation.
\end{abstract}

\section{Introduction}
\label{sec:intro}

LLM-based agents are increasingly deployed for enterprise data intelligence (DI), answering business questions that require retrieving knowledge, querying relational databases, and applying business-specific rules. Consider a business analyst asking: ``What percentage of customer-support tickets met their resolution-time commitment last month?'' A naive agent architecture is not going to be sufficient because simply subtracting each ticket's opening time from its closing time is not enough. It must first retrieve the applicable service-level rules: premium customers have shorter resolution targets, the clock pauses while a ticket is waiting for the customer, and weekends are excluded for plans without 24-hour support. The agent must then query ticket histories and customer-plan data, apply the correct rule to each ticket and compute eligible resolution time. Answering the question therefore requires combining retrieved business rules with analytical data and computation; ignoring the rules would produce an incorrect result.

\begin{table*}[!htbp]
\centering
\resizebox{\textwidth}{!}{
\newcommand{\yes}{\ding{51}}
\newcommand{\no}{\ding{55}}
\begin{tabular}{p{3.3cm} C{1.8cm} C{1.8cm} C{3.9cm} C{3.4cm} C{3cm} C{2.3cm} C{1.8cm} C{1.5cm}}
\toprule
& \multicolumn{6}{c}{\textbf{Task Types}} & \multicolumn{2}{c}{\textbf{Construction Method}} \\
\cmidrule(lr){2-7} \cmidrule(lr){8-9}
& \makecell{SQL\\Querying}
& \makecell{Document\\QA}
& \makecell{Numeric Computation\\\footnotesize(e.g., point query, comparison)}
& \makecell{Analytical Reasoning\\\footnotesize(e.g., trend, stats)}
& \makecell{Document\\ Retrieval}
& \makecell{Document\\ Alters Answer}
& \makecell{Q\&A\\Generation}
& \makecell{Quality\\Check} \\
\midrule
Spider                              & \yes      & \no  & \yes & \no  & \no & \no   & M     & M     \\
BIRD                                & \yes      & \no  & \yes & \no  & \no &  \no   & M     & M     \\
Spider 2.0                          & \yes      & \no  & \yes & \no  & \no & \yes  & M>L   & M     \\
$\tau$-bench                        & \no  & \yes & \yes & \no  & \no & \yes  & M>L   & M     \\
InfiAgent-DABench$^{\ddagger}$      & \no    & \no  & \yes & \yes & \no & \no   & L>M   & M, L   \\
FinanceBench                        & \no       & \yes & \yes & \no  & \yes & \yes   & T>M   & M     \\
AutoBencher$^{\dagger}$             & \no       & \yes & \yes & \no  & \no & \no   & L     & M, L   \\
\textbf{DI-Bench}$^{\ddagger}$      & \yes      & \yes & \yes & \yes & \yes & \yes  & G>T>L & M, L   \\
\bottomrule
\end{tabular}
}
\caption{Comparison of DI-Bench against benchmarks for analytical / tool-using agents. Symbols on benchmark names indicate the released artifact: $\dagger$ pipeline only; $\ddagger$ both pipeline and fixed dataset; no symbol means dataset only. The \emph{document retrieval} column indicates whether the agent must retrieve the relevant document. The \emph{document alters answer} column specifies if the answer correctness relies on the document. The pair \ding{55}\,/\,\ding{51}\ for these two columns (no retrieval, document needed) appears when the document affects the answer but is given to the agent up front, rather than retrieved. Construction method: M = manual, L = LLM, T = template, G = graph; ``$>$'' orders components by contribution.
}
\label{tab:benchmark_comparison}
\end{table*}
 Evaluating agents on this sequence of tasks from knowledge retrieval, data querying, business rule understanding and downstream analytical tasks demands benchmarks that test this full stack, but existing benchmarks each cover only part of it. Text-to-SQL benchmarks like Spider~\citep{yu2018spider} and BIRD~\citep{li2023bird} provide the question and any required knowledge inline, so agents never retrieve from a separate corpus. Spider~2.0~\citep{lei2025spider} uses per-task documents, which specify formulas or thresholds the gold SQL implements, but each task names the document it needs, and the documents are authored for that one task rather than as a corpus the agent must navigate. Data-analysis benchmarks such as InfiAgent-DABench~\citep{hu2024infiagent} broaden task coverage but operate on a single 
  data table per task with no document corpus. Document-grounded financial QA benchmarks~\citep{islam2023financebench, chen2021finqa} require retrieving from filings, but these documents are data sources for numerical values rather than natural language rules that govern computation over a database. Tool-use benchmarks like $\tau$-bench~\citep{yao2024tau} test rule-following on API calls, with rules given in-context and no analytical computation over data. Concurrent data-science and enterprise-data-agent benchmarks~\citep{jing2025dsbench, sahu2025insightbench, ma2026can, yang2026aidabench, abaskohi2025drbench, mo2026entworld} address different task types, including modeling, insight discovery, cross-system integration, end-to-end analytics, deep research, and GUI pipeline completion, rather than retrieving and applying business knowledge during analysis. Table~\ref{tab:benchmark_comparison} summarizes these benchmarks.

A second line of work addresses the cost of manual benchmark construction. LLM-driven generation~\citep{li2025autobencher} constructs both questions and answers with an LLM, providing no formal guarantee of answer correctness. In multi-hop text QA, \citet{min2019compositional} showed that nominally multi-hop benchmarks were often answerable from a single hop, motivating generation pipelines that structurally guarantee multi-hop necessity. 2WikiMultiHopQA~\citep{ho2020constructing} composes questions from Wikidata triples via typed graph traversal, while MuSiQue~\citep{trivedi2022musique} uses counterfactual checks to verify that each evidence hop is required. These multi-hop text QA efforts, however, operate over unstructured text only: their graphs link entities from a knowledge corpus rather than database tables and rule documents, and verification targets evidence necessity rather than the value of a computed answer. The latter matters for analytical questions, where a document can be cited yet leave the numeric answer unchanged. While recent progress in retrieval and text-to-SQL has substantially improved individual capabilities, it remains unclear whether agents can correctly integrate retrieved business knowledge into downstream analytical computation. Moreover, enterprise deployments rely on organization-specific metrics and business logic that evolve over time and differ substantially across domains. 
A benchmark built for one company does not transfer to another, and even within one company it goes stale as definitions are revised. 
A team deploying an agent therefore needs to generate a benchmark from its own database and business documents, update the benchmark as metrics and rules change, rather than reuse a static, general-purpose test set. This motivates a benchmark generation pipeline that any organization can apply to its own domain.

We address this gap with DI-Bench, an end-to-end pipeline for generating verifiable benchmarks for data intelligence agents operating over both relational databases and unstructured business knowledge. By combining programmatic execution, automated validation, and realistic enterprise artifacts, DI-Bench produces benchmark tasks with deterministic ground truth while reducing manual benchmark construction effort.

Our contributions are
\textbf{1. DI-Bench}, a pipeline for generating data intelligence benchmarks from structured databases and business knowledge sources.
\textbf{2. Demonstration on a generated benchmark} spanning Brazilian e-commerce and Czech retail banking, producing 731 validated tasks across nine analysis types that require retrieval, computation, and business-rule reasoning.
\textbf{3. An empirical analysis of enterprise agents.} Across models, DI-Bench reveals consistent performance degradation on tasks requiring integrating retrieved business rules with analytical computation.

\section{Problem Setting}

 DI-Bench targets questions such as the customer-support example from Section~\ref{sec:intro}: ``What percentage of customer-support tickets met their resolution-time commitment last month?'' This percentage cannot be computed correctly from ticket timestamps alone. The agent must query ticket histories and customer-plan data, retrieve the applicable service-level rules, and apply those rules before aggregating the result. More generally, DI-Bench studies settings in which answering a business question requires both query execution over relational data and retrieval-augmented generation over documents containing computation-relevant rules~\citep{lewis2020retrieval,guu2020retrieval}. We assume two inputs:
  \begin{enumerate}
      \setlength{\itemsep}{2pt}
      \setlength{\parskip}{0pt}
      \item A \textit{relational database} with a known schema, including tables, columns, and foreign keys. In the running example, the database
      contains ticket histories and customer service plans.
      \item A \textit{knowledge base} (KB) of \textit{rule documents} containing business definitions, thresholds, and conditional overrides that
      affect a metric's computation. In the running example, the applicable resolution-time target may depend on the customer's service plan and weekend-exclusion rule. The KB also includes a metric
      catalog with formulas and sample SQL templates. In our demonstration, these documents are LLM-generated augmentations grounded in the database
      schema (Appendix~\ref{app:rule_provenance}).
  \end{enumerate}

The critical property is information scattering: no single artifact contains everything needed to answer a question correctly. For example, the metric catalog defines Customer Satisfaction (CSAT) as the share of reviews scoring 4 or 5 stars, but a separate rule document specifies that neutral (3-star) reviews must be excluded from the denominator. Without retrieving that rule, the computed answer deviates by several percentage points, making each task genuinely discriminative.

\section{Methodology}
\label{sec:method}

\begin{figure*}[!ht]
  \centering
  \includegraphics[width=0.9\textwidth, height=8cm, keepaspectratio=false]{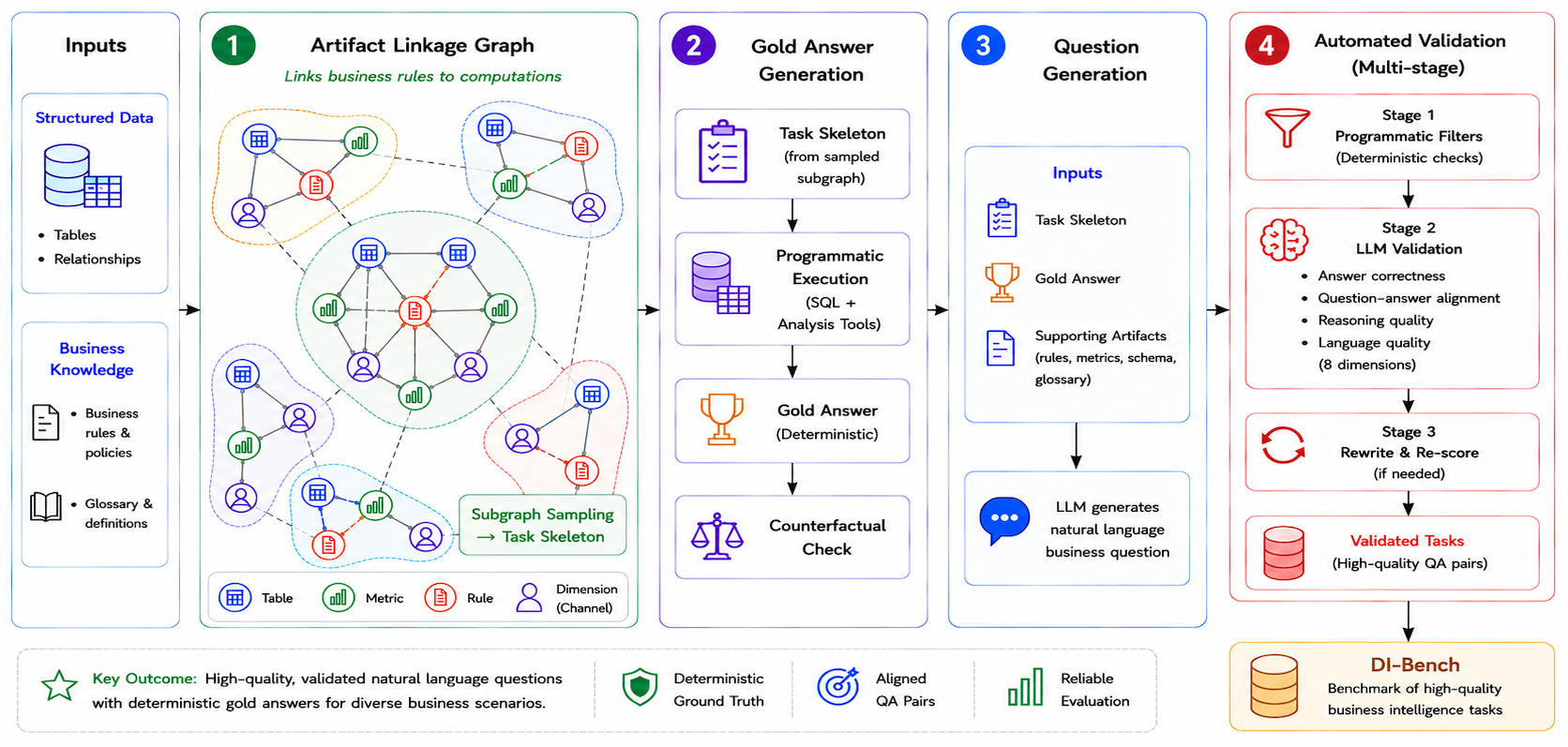}
  \caption{The DI-Bench pipeline. From a database and knowledge base, the pipeline (1) constructs an artifact linkage graph connecting metrics to required tables and governing rule documents, sampling connected subgraphs as task skeletons, (2) computes gold answers via query execution, (3) generates natural language questions backward from the answers, and (4) applies automated validation filters to verify each task. } 
  \label{fig:pipeline}
\end{figure*}

DI-Bench generates benchmark tasks through five stages: (1) construct a graph that links structured data to the unstructured documents that affect its interpretation, (2) sample subgraphs from the graph as task skeletons, (3) execute SQL and tool(s) to compute gold answers, (4) generate natural language questions backward from those answers, and (5) validate quality. Figure~\ref{fig:pipeline} provides an overview.

\subsection{Artifact Linkage Graph}
\label{sec:alg}

To guarantee that each task genuinely tests document retrieval, the dependencies between documents and computations are formalized as a typed \emph{artifact linkage graph} (ALG). Such tasks that involve both document retrieval and analytical computations are common in real enterprise tasks, but current benchmarks lack reliable methods to generate them. The graph immediately shows which unstructured documents map to which structured tables and metrics, allowing us to generate tasks that meaningfully involve both skills.

\paragraph{Graph Structure.}
The ALG contains four node types: tables, metrics, rules, and dimensions. Six edge types capture dependencies between structured data and business knowledge (Table~\ref{tab:edge_types}). The key edge is \emph{rule$\rightarrow$metric}, which records that a document modifies the computation of a metric.

\begin{table}[t]
\centering
\resizebox{1.00\columnwidth}{!}{
\begin{tabular}{@{}lp{5.0cm}@{}}
\toprule
\textbf{Edge} & \textbf{Meaning} \\
\midrule
metric $\to$ table & Metric formula uses columns from table \\
table $\to$ dimension & Table has column usable as a filter/grouping dimension \\
table $\to$ table & Tables can be joined \\
rule $\to$ metric & Rule document changes this metric's computed answer \\
metric $\to$ metric & Composite metric uses another as a component \\
rule $\to$ dimension & Rule applies only when a dimension takes a specific value\\
\bottomrule
\end{tabular}}
\caption{Edge types in the artifact linkage graph. The rule$\to$metric edge is key: it connects unstructured documents to the structured computations they modify.}
\label{tab:edge_types}
\end{table}

\paragraph{Construction.}
Edges among structured artifacts are extracted directly from schemas and metric definitions. Rule-related edges require interpreting natural language documents; we use an LLM (DeepSeek V3.2) to identify which metrics a document modifies and whether it applies conditionally to specific dimension values.

\paragraph{Imperfect Graph Extraction.}
The pipeline does not assume perfect rule--metric extraction. Candidate links are evaluated through counterfactual checks during benchmark generation. Counterfactual checks assess whether the inclusion of the rule changes the SQL or answer. These checks help filter spurious rule--metric associations and improve benchmark quality. Imperfect extraction therefore primarily affects coverage rather than correctness (Appendix~\ref{app:kg_stats}).

\paragraph{Why a graph?}
The graph captures dependencies between business documents and the computations they influence, enabling systematic generation of tasks that genuinely require both data access and business knowledge, rather than tasks where a document is nominally attached but does not affect the answer. 
Because a rule$\rightarrow$metric edge is defined by the rule's causal effect on the computation rather than surface simil
arity in wording, the graph selects answer-changing rules far more reliably than semantic-similarity or random pairing (See analysis in Appendix~\ref{app:ablation_ALG}).

\subsection{Subgraph Sampling}
\label{sec:sampling}

The pipeline samples connected subgraphs from the ALG. Each subgraph provides a task skeleton that fixes the metric, data sources, dimensions, and business rules, but not their concrete values. Each skeleton yields multiple benchmark tasks by varying filters, time windows, and analysis types.

\paragraph{Enumeration.}
Starting from each metric node, the pipeline expands along graph edges to collect required tables, dimensions, and modifying rule documents. Valid connected subgraphs are then enumerated or sampled depending on dataset size and target task count, while respecting rule applicability and metric dependencies.

\paragraph{Selection for quality and diversity.}
From the pool of valid candidate subgraphs, the pipeline selects the subset that will be instantiated as tasks, aiming for a set that is
diverse in the metrics, rules, and dimensions it covers rather than one that repeats similar combinations. Selection uses Maximal Marginal Relevance \citep[MMR;][]{carbonell1998use}, which scores each candidate on semantic coherence, rule coverage, and join complexity, then greedily balancing quality against distance from already-selected subgraphs. A post-selection pass ensures every metric and rule document appears at least once (Appendix~\ref{app:sampling_details}). 

\paragraph{Minimum coverage augmentation.} The target minimum per analysis type is configurable. After the initial generation round, the pipeline identifies types below the target, samples additional compatible subgraphs, and reruns question generation, SQL execution, answer enrichment, and validation. A candidate is retained only if its gold answer can be reproduced by re-executing its stored SQL. In our demonstration, this augmentation adds 23 tasks and raises counterfactual, decomposition, and significance testing to at least 10 tasks in each domain.

\paragraph{Example} In the running example from Section~\ref{sec:intro}, the sampled subgraph would involve the metric for ticket resolution-time; the connected rules for premium customers, the clock pausing rule, and weekend exclusion; the tables containing ticket histories and customer service plans; and the resolution-date dimension used to restrict the computation to the previous month.

\subsection{Ground Truth Computation}
\label{sec:ground_truth}

Since an agent is scored against the gold answer, a reliable benchmark depends on that answer being correct. Rather than have a model produce it, DI-Bench computes each gold answer by executing SQL against the real database, making it deterministic and correct by construction.
Each subgraph, after fixing dimension values and the analysis type, yields concrete tasks whose gold answers are computed using the following two steps:

\paragraph{Step 1: SQL parameterization.} Each metric in the catalog has a SQL template with placeholders for dimension filters (e.g., \texttt{state}, \texttt{start\_date}), which are substituted with concrete values drawn from the data. Each rule document is associated with a structured SQL modification, such as an additional \texttt{WHERE} or \texttt{HAVING} clause, that encodes its effect on metric computation. These modifications are generated by LLM with human expert verification. These modifications are applied to the metric's base SQL, and the gold answer is produced by executing the query on database. %

\paragraph{Step 2: Analysis-type enrichment.} Depending on the assigned analysis type (e.g., trend, comparison, ranking), a post-processing step transforms the raw SQL result into a structured gold answer. For instance, a trend task runs the query over consecutive periods to identify direction, while a comparison task computes a $p$-value from two group distributions. Nine analysis types are supported (Appendix~\ref{app:task_types}), each drawn from a compatibility set determined by the subgraph's dimension types, where temporal dimensions support trend analysis and spatial or categorical dimensions support comparison and ranking. In the business analyst example from Section~\ref{sec:intro}, we might add on the task of comparing the percentage of customer-support tickets meeting their resolution-time commitment last month to the month before that (a comparison analysis type).

\subsection{Backward Question Generation}

With the gold answer computed, an LLM (DeepSeek V3.2) generates a natural language question that would elicit that answer, following the answer-conditioned question-generation paradigm~\citep{du2017learning}. The prompt provides the metric name, gold answer, analysis type, and required rule documents. The model produces a question resembling what a business stakeholder would ask, without exposing database internals or hinting at which documents to consult. Because the answer is already known to be correct, the LLM's only job is natural phrasing and poor phrasing is caught by validation.

\subsection{Automated Validation}
\label{sec:validation}

Prior to inclusion in the benchmark, each generated question-answer pair is subjected to a two-stage validation pipeline: 

\paragraph{Stage 1: Programmatic Filters.} The first stage applies deterministic, regex-based checks for structural defects and content-level issues without LLM inference. Checks cover structural integrity (e.g., flagging leakage of internal identifiers and degenerate short questions), dimension sanity (e.g., rejecting tasks where a datetime field was mistakenly treated as a categorical segmentation), answer completeness (e.g., verifying that a valid, scorable gold answer is present), and column name naturalization (e.g., replacing schema identifiers with natural language). A task passes this stage only if zero issues are detected.

\paragraph{Stage 2: LLM Quality Scoring.}
Candidates that clear the programmatic stage are then scored by an LLM judge~\citep{zheng2023judging, liu2023g} (DeepSeek V3.2) across eight binary quality dimensions covering objectivity, leakage, naturalness, relevance, format alignment, correctness, reasoning, and scope consistency. 
A task is admitted only if all eight dimensions pass.
Tasks failing only on phrasing-related dimensions are automatically rewritten and re-scored before final filtering. 
In Appendix~\ref{app:validation}, we report the audit against human review, which shows that admitted tasks from the LLM judge achieve 99\% average pass rate across all dimensions on evaluated datasets.

\section{Experiments}
\label{sec:experiments}

\paragraph{Demonstration datasets.} We apply DI-Bench on two public relational datasets spanning distinct industries. The first is the Olist Brazilian E-Commerce dataset~\citep{olist2018brazilian} consisting of 9 tables (orders, items, payments, reviews, products, sellers, customers, and geolocation) covering roughly 100K orders across all 27 Brazilian states. The second is the PKDD'99 Czech financial dataset~\citep{berka2000guide}, modeled as the bank \emph{\v{C}esk\'{a} Banka}: 8 tables (accounts, clients, dispositions, transactions, orders, loans, cards, and districts) capturing retail banking activity. It stores dates as integers and distributes data across more tables requiring additional joins, both of which increase task difficulty. For each domain we construct a knowledge base of 29 metric definitions and 17 business-rule documents (thresholds, exclusions, weighting formulas, and conditional overrides), from which DI-Bench builds an artifact linkage graph of 71--76 nodes and 168--191 edges (Appendix~\ref{app:kg_stats}). The two domains differ in schema shape, terminology, and rule structure, testing the pipeline's transferability. All documents and metrics are fictional and do not represent any real organization.

\paragraph{Benchmark composition.} DI-Bench contains 731 tasks: 449 from Brazilian e-commerce and 282 from Czech retail banking. 
These tasks passed the automated validation, which rejects 21\% of candidates at Stage 1 and 26\% of the remainder at Stage 2.
Minimum-coverage augmentation contributes 23 of these tasks, including 4 Brazilian decomposition tasks and 19 Czech tasks (2 counterfactual, 9 decomposition, and 8 significance-testing). The augmentation raises counterfactual, decomposition, and significance testing to at least 10 instances per domain. All aggregate and per-type results below use the complete 731-task benchmark.

\subsection{Setup}

To show that DI-Bench is challenging and can discriminate among model capabilities, we evaluate agents using various models on it. Agents are equipped with tools for SQL execution and knowledge base retrieval. The database schema is provided in the system prompt. Agents interact through the Bedrock Converse API with native tool use and a maximum of 20 turns per task. We evaluate Claude Opus 4.6, Claude Haiku 4.5, DeepSeek V3.2, and Qwen3-80B, with temperature set to 0 and maximum output tokens of 4096. Scoring uses exact match with task-specific tolerance rules. 

\subsection{Model Performance on DI-Bench}

\begin{table}[t]
\centering
\small
\begin{tabular}{@{}lcccc@{}}
\toprule
\textbf{Model} & \textbf{BR} & \textbf{CZ} & \textbf{All} & \textbf{Turns} \\
\midrule
Claude Opus 4.6 & \textbf{59.2} & \textbf{55.7} & \textbf{57.9} & 5.5 \\
DeepSeek V3.2 & 56.1 & 42.6 & 50.9 & 11.5 \\
Claude Haiku 4.5 & 52.8 & 42.9 & 49.0 & 5.5 \\
Qwen3-80B & 53.0 & 39.7 & 47.9 & 5.2 \\
\bottomrule
\end{tabular}
\caption{Agent accuracy (\%) and efficiency on DI-Bench. BR = Brazilian E-Commerce (449 tasks), CZ = Czech Financial (282 tasks), 731 total. Avg.\ Turns = mean LLM calls per task (max 20).}

\label{tab:main_results}
\end{table}
\begin{figure*}[t]
  \centering
  \includegraphics[width=0.83\textwidth]{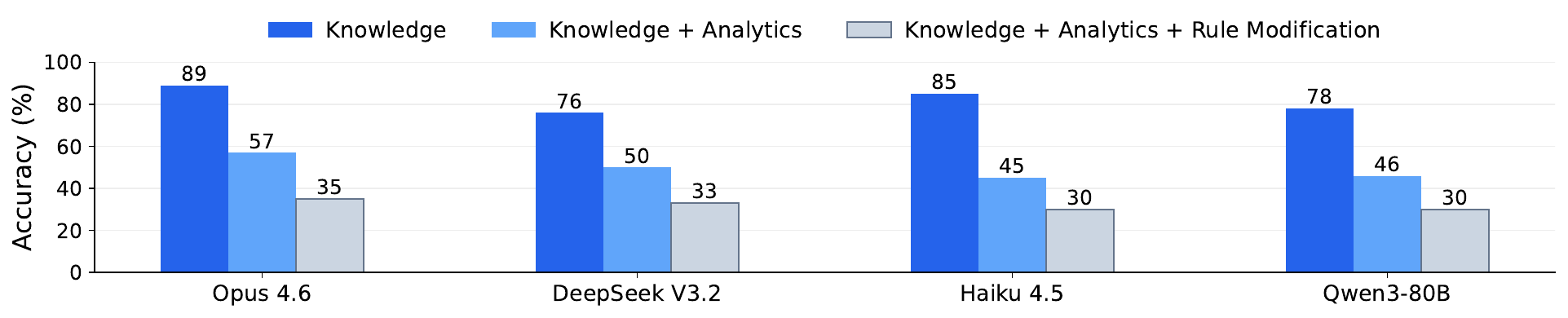}
  \caption{
Knowledge vs. Knowledge-grounded analytical tasks across models. Models score highest on \emph{Knowledge} tasks, lower once \emph{Analytics} is added, and lowest when a retrieved rule must further modify the computation. The resulting performance drop reveals a persistent knowledge-to-computation gap: retrieving business knowledge is largely solved, but correctly applying that knowledge during computation remains a major challenge.
  }
  \label{fig:integration_gap}
\end{figure*}

\paragraph{Analysis.} Opus achieves the highest overall accuracy (57.9\%), followed by DeepSeek V3.2 (50.9\%), Haiku (49.0\%), and Qwen3-80B (47.9\%). The gap between BR and CZ remains approximately 10 percentage points across models, reflecting the Czech dataset's less familiar banking terminology. DeepSeek requires more interaction steps (11.5 turns on average versus approximately 5 for the other models), indicating lower tool-use efficiency.

\paragraph{Knowledge-to-Computation Gap.}
We partition tasks into three incremental groups (Figure~\ref{fig:integration_gap}): \emph{Knowledge} (knowledge-base questions answered through retrieval alone, $N=137$); \emph{Knowledge + Analytics} (retrieve a metric definition and compute it over the database, $N=419$); and \emph{Knowledge + Analytics + Rule Modification} (additionally retrieve and apply a rule that modifies how the metric is computed, $N=175$). These task types are generated from different subgraph structures.

Averaged across the four models, accuracy drops from 82\% on
Knowledge tasks to 50\% on Knowledge + Analytics and further to
32\% on Knowledge + Analytics + Rule. This decline reveals a substantial knowledge-to-computation gap: all four models can often retrieve the relevant business knowledge, yet struggle to correctly incorporate it into downstream computation. The trend is consistent across all evaluated models, indicating a systematic limitation that is not captured by retrieval-only or text-to-SQL benchmarks. Section~\ref{sec:ablation_robustness} uses matched variants to isolate rule application.

\paragraph{Task type difficulty.}
Accuracy varies by analysis type on the 731-task
benchmark (Table~\ref{tab:by_type}). Ranking and point query remain
difficult because every row in the result must match exactly.
Minimum-coverage augmentation raises counterfactual, decomposition,
and significance testing to at least 10 tasks per domain; these
results remain descriptive because the cells are still small. In contrast, tasks requiring only a directional or binary judgment (threshold, trend, comparison) score 43--81\%.

\begin{table}[t]
\centering
\small
\resizebox{0.95\columnwidth}{!}{
\begin{tabular}{lrcrrc}
\toprule
\textbf{Task Type} & \multicolumn{2}{c}{\textbf{BR (449)}} & & \multicolumn{2}{c}{\textbf{CZ (282)}} \\
\cmidrule{2-3} \cmidrule{5-6}
 & \textbf{N} & \textbf{Acc} & & \textbf{N} & \textbf{Acc} \\
\midrule
Ranking & 92 & 14 & & 53 & 29 \\
Knowledge retrieval & 69 & 87 & & 68 & 78 \\
Threshold crossing & 81 & 78 & & 42 & 43 \\
Point query & 54 & 28 & & 48 & 20 \\
Trend analysis & 70 & 78 & & 8 & 47 \\
Comparison & 30 & 81 & & 33 & 55 \\
Significance testing & 32 & 58 & & 10 & 80 \\
Counterfactual & 11 & 5 & & 10 & 10 \\
Decomposition & 10 & 0 & & 10 & 10 \\
\bottomrule
\end{tabular}}
\caption{Task volume and accuracy ($\%$) by types, averaged across the four models. Knowledge retrieval, comparison, threshold, and trend are easier (binary or single-fact match); point query and ranking require exact match across multiple rows; counterfactual and decomposition are hardest, requiring multi-rule reasoning.}
\label{tab:by_type}
\end{table}

\subsection{Failure Analysis}

Manual inspection of Claude Opus 4.6 errors reveals two dominant failure modes. In the first, the agent retrieves the relevant metric but never incorporates the associated rule knowledge into its SQL, as seen on a query requiring the on-time-delivery SLA where the agent omitted the one-day grace period, producing systematically incorrect defect counts. In the second, the agent retrieves and acknowledges the correct rule in its reasoning yet executes SQL identical to the naive version, leaving the answer unchanged. Both modes confirm that the benchmark's primary signal is not whether the agent can locate the relevant rule but whether it correctly applies it during computation, with additional failure cases discussed in Appendix~\ref{app:failure_cases}.

\section{Ablation and Robustness Analyses}
\label{sec:ablation_robustness}

\begin{table}[t]
\centering
\small
\setlength{\tabcolsep}{4pt}
\begin{tabular}{@{}lccc@{}}
\toprule
Model & Rule & Oracle & No rule \\
\midrule
Opus 4.6 & 35.0 & 37.3 & 53.6 \\
DeepSeek V3.2 & 34.0 & 31.4 & 54.3 \\
Haiku 4.5 & 30.0 & 29.4 & 49.3 \\
Qwen3-80B & 31.0 & 32.7 & 54.3 \\
\midrule
Average & \textbf{32.5} & \textbf{32.7} & \textbf{52.9} \\
\bottomrule
\end{tabular}
\caption{Accuracy (\%) on 175 rule-grounded tasks. Task content and answer format are fixed across conditions.}
\label{tab:matched_rule_oracle}
\end{table}

\paragraph{Rule application.} We construct matched variants for 175 rule-grounded tasks, holding the metric, dimension, time window, analysis type, and answer format fixed. The results are shown in Table~\ref{tab:matched_rule_oracle}. The standard condition requires retrieving and applying the rule, the oracle condition provides the rule directly, and the no-rule condition uses the corresponding standard metric. Oracle access changes average accuracy by only $+0.2$ points (32.5\% to 32.7\%), whereas removing the rule requirement raises accuracy by 20.4 points (32.5\% to 52.9\%). Within \texttt{threshold\_crossing} ($N=35$), accuracy likewise rises from 62.1\% to 77.9\% without the rule, indicating that rule application rather than retrieval is the primary bottleneck.

\paragraph{Generator and LLM judge sensitivity.} On 87 valid paired examples with the same task skeletons, Qwen3-80B and DeepSeek-generated questions receive similar mean quality scores under an independent Claude Sonnet 4.6 LLM judge (79.9\% versus 79.6\%). Across four validator judges, precision is 94--96\% and recall is 79--93\% against the same balanced human-labeled sample. These checks bound but do not eliminate the shared-model confound; Appendix~\ref{app:generator_judge} provides details.

\section{Conclusion}

We presented DI-Bench, a transferable pipeline for generating in-domain benchmarks that evaluate agents on data-intelligence tasks requiring both knowledge retrieval and analytical computation. By linking structured databases with unstructured business documents and computing gold answers through programmatic execution, DI-Bench produces realistic benchmark tasks with deterministic ground truth. Demonstrated on two public domains, the resulting 731-task benchmark reveals a substantial knowledge-to-computation gap in all four evaluated models: while retrieval-only tasks reach 82\% accuracy, performance drops to 32\% when retrieved business knowledge must be incorporated into computation. These findings suggest that retrieving business knowledge is not the primary challenge for enterprise agents, but correctly grounding it in downstream tasks remains a major bottleneck. More broadly, DI-Bench provides a scalable framework for evaluating agentic capability as business logic, metrics, and data evolve.

\section*{Limitations}

The pipeline is demonstrated on two domains (e-commerce and banking); broader validation across additional industries would strengthen the transferability claim. Future work could evaluate the benchmark under alternative retrieval systems and more complex agent architectures.

The business-rule documents and metric catalogs used in our demonstration are synthetic augmentations rather than artifacts from a real enterprise. We made this design choice because we are not aware of any publicly available dataset that simultaneously provides relational data, business-rule documentation, and permission for open benchmark release. The generated artifacts are grounded in the underlying schemas and are designed to simulate realistic business rules, metric definitions, thresholds, exclusions, and conditional overrides. Consequently, the benchmark is intended to approximate realistic enterprise analytical pipelines rather than reproduce the exact business processes of any specific organization. Nevertheless, the generated artifacts may not fully capture the ambiguity, inconsistency, evolving governance, and cross-team dependencies commonly present in production business documentation, which motivates evaluation on more realistic enterprise documentation. 

Additionally, scoring is based on exact match and does not award partial credit for correct reasoning with minor computational errors.
Finally, although the 731-task benchmark includes at least 10 counterfactual, decomposition, and significance-testing tasks per domain ($N{=}21/20/42$ overall), these cells remain too small for strong inferential comparisons by analysis type.

\bibliography{custom}
\clearpage

\appendix
\section*{Appendix}
\section{Analysis Type Definitions}
\label{app:task_types}

Each analysis type corresponds to a composition of primitive operators (Retrieve, Filter, Aggregate, Compare, Rank, Temporal, Test). Table~\ref{tab:app_types} defines the 9 types used in DI-Bench.

\begin{table*}[ht]
\centering
\small
\begin{tabular}{@{}l l p{4.6cm} p{4.4cm}@{}}
\toprule
\textbf{Type} & \textbf{Operators} & \textbf{Example Question} & \textbf{Gold Answer (example)} \\
\midrule
Knowledge retrieval & Ret & What constitutes an outlier that should be excluded when calculating the average delivery time? & {\scriptsize\ttfamily ``Delivery duration exceeding sixty (60) calendar days is excluded as a data anomaly.''} \\
Point query & Ret\,$\rightarrow$\,Fil\,$\rightarrow$\,Agg & For our Q2 2018 performance review, what percentage of transactions were paid with credit cards? & {\scriptsize\ttfamily [\{payment\_type: credit\_card, pct: 80.08\}]} \\
Trend analysis & Ret\,$\rightarrow$\,Tmp\,$\rightarrow$\,Agg & Show me the monthly trend of our average delivery time throughout 2017. & {\scriptsize\ttfamily \{direction: increasing, total\_change\_pct: 32.1, peak\_period: 2018-01\}} \\
Comparison & Ret\,$\rightarrow$\,Fil\,$\rightarrow$\,Agg\,$\rightarrow$\,Cmp & Which of these two states --- PR or AL --- had a faster average delivery time for shipped and delivered orders during Q3 2017? & {\scriptsize\ttfamily \{group\_a: PR, group\_b: AL, value\_a: 10.7, value\_b: 28.52, higher: AL\}} \\
Ranking & Ret\,$\rightarrow$\,Fil\,$\rightarrow$\,Agg\,$\rightarrow$\,Rnk & Which 5 states have the highest number of active sellers? & {\scriptsize\ttfamily [\{SP: 1638\}, \{PR: 316\}, \{MG: 220\}, \{SC: 172\}, \{RJ: 175\}]} \\
Significance testing & Ret\,$\rightarrow$\,Fil\,$\rightarrow$\,Agg\,$\rightarrow$\,Tst & Are the average installment counts for orders paid by credit card and boleto in the first quarter of 2018 statistically significantly different from each other? & {\scriptsize\ttfamily \{significant: true, p\_value: 0.0, test\_name: Welch's t-test (two-tailed)\}} \\
Threshold crossing & Ret\,$\rightarrow$\,Fil\,$\rightarrow$\,Agg\,$\rightarrow$\,Cmp & Did the average delivery time for customers in Acre exceed 20.0 days? & {\scriptsize\ttfamily \{value: 19.79, threshold: 20.0, result: FAIL\}} \\
Counterfactual & Ret\,$\rightarrow$\,Fil\,$\rightarrow$\,Agg\,$\rightarrow$\,Cmp & What's the average revenue we generated per unique customer in S\~{a}o Paulo state, assuming we could extend our operations there starting from January 2017? & {\scriptsize\ttfamily [\{customer\_state: SP, revenue\_per\_customer: 129.41\}]} \\
Decomposition & Ret\,$\rightarrow$\,Fil\,$\rightarrow$\,Agg\,$\rightarrow$\,Grp & Can you break down the Order Defect Rate by seller state for the first half of 2018, showing which geographic areas are contributing the most to our overall defect rate? & {\scriptsize\ttfamily [\{BA: 18.18\}, \{CE: 0.0\}, \{DF: 13.21\}, \{ES: 18.18\}, \ldots]} \\
\bottomrule
\end{tabular}
\caption{Analysis type definitions with operator compositions, example questions, and a representative gold answer (Ret = Retrieve, Fil = Filter, Agg = Aggregate, Tmp = Temporal, Cmp = Compare, Rnk = Rank, Tst = Test, Grp = Group). Gold answers range from a single value or pass/fail verdict to multi-row breakdowns that must match exactly, which explains the accuracy ordering in Table~\ref{tab:by_type}.}
\label{tab:app_types}
\end{table*}

\section{Knowledge Base Examples}
\label{app:kb_examples}

\paragraph{Rule document (excerpt).} The following is a shortened version of the Active Seller Definition Rule, an example rule document:

\begin{quote}
\small
\textbf{Rule Title:} Active Seller Definition Rule \\
\textbf{Effective Date:} 1 July 2018 \\
\textbf{Core Rule:} A seller is classified as Active if and only if both conditions are met within a trailing 90-day window: (1) Volume: at least one delivered order; (2) Quality: average review score $\geq$ 3.0 across all historical reviews. Sellers meeting volume but not quality are classified as ``On Probation'' and excluded from Active Seller counts.
\end{quote}

\paragraph{Metric catalog entry (excerpt).}

\begin{quote}
\small
\textbf{MV-SEL-001: Active Seller Count} (Tier L2) \\
\textbf{Formula:} COUNT(DISTINCT seller\_id) WHERE delivered AND avg\_review $\geq$ 3.0 \\
\textbf{Tables:} order\_items, orders, order\_reviews 
\end{quote}

\subsection{Rule Document Provenance}
\label{app:rule_provenance}

The rule documents and metric catalogs used in our demonstration are LLM-generated augmentations, not real enterprise documents. Given that the two underlying datasets are public databases with no accompanying rule documentation, we augment the datasets with realistic rule documents.

\paragraph{Generation process.} For each domain, an LLM (DeepSeek V3.2) was prompted with the database schema, a metrics catalog (also LLM-generated from the schema), and supporting context about the fictional company. The LLM was instructed to produce standalone rule documents whose rules would change the computed value of specific metrics if not consulted. Each document was generated independently in a single LLM call.

\paragraph{Generation prompt (metric rules, abbreviated).}

\begin{tcolorbox}[colback=gray!3, colframe=gray!50!black, title={\textbf{Rule Document Generation Prompt (abbreviated)}}, fonttitle=\small, breakable]
\small\ttfamily
\rmfamily\textbf{System:} You are a business analyst writing internal rule documents for \{company\}.\\[4pt]
\ttfamily
Write a standalone rule document that defines rules which modify how business metrics are computed. The rule should contain information NOT already in the metrics catalog. The goal is that someone computing the metric would get a WRONG answer if they don't consult this rule document. Include: a realistic effective date, named owner, version, core rules/thresholds, scope (affected metrics), exceptions, and a rationale section.\\[6pt]
\rmfamily\textbf{Context provided:}\\
--- Metrics catalog ---\\
--- Supporting documents ---\\
\end{tcolorbox}

\paragraph{Descriptive statistics.}
Each domain has 17 rule documents, averaging 750--820 words per document (Table~\ref{tab:rule_doc_stats}). Each document falls into one of six rule types that govern how it modifies a metric: \emph{definitional} (defines what counts, e.g., active seller criteria), \emph{exclusionary} (removes rows, e.g., exclude neutral reviews), \emph{temporal/versioned} (rule changes on a date, e.g., SLA change in January 2018), \emph{conditional} (applies only to a dimension slice, e.g., S\~{a}o Paulo orders), \emph{privacy guardrail} (suppresses small-$n$ results), and \emph{override} (replaces a default formula under specific conditions). All six types are represented in both domains.

\begin{table}[h]
\centering
\small
\begin{tabular}{lcc}
\toprule
& \textbf{Brazilian} & \textbf{Czech} \\
\midrule
Number of rule documents & 17 & 17 \\
Mean word count & 821 & 756 \\
Min / max word count & 654 / 1132 & 592 / 928 \\
Median word count & 847 & 721 \\
Rule types represented & 6  & 6  \\
\bottomrule
\end{tabular}
\caption{Descriptive statistics for the LLM-generated rule documents. ``Rule types represented'' shows the count out of 6 possible types (definitional, exclusionary, temporal, conditional, privacy, override).}
\label{tab:rule_doc_stats}
\end{table}

\paragraph{Example rule document (excerpt).} The following is an abbreviated version of the Outlier Exclusion Rule from the Brazilian e-commerce domain:

\begin{tcolorbox}[colback=gray!3, colframe=gray!50!black, title={\textbf{Outlier Exclusion Rule for Delivery Time Metrics (excerpt)}}, fonttitle=\small, breakable]
\small
\textbf{Rule Title:} Outlier Exclusion Rule for Delivery Time Metrics\\
\textbf{Effective Date:} October 1, 2018\\
\textbf{Owner:} Ana Silva, Head of Logistics Analytics\\[4pt]
\textbf{Core Rule:} Any delivered order with a calculated delivery duration exceeding 60 calendar days is classified as a data anomaly and excluded from metric computation.\\[4pt]
\textbf{Scope:} On-Time Delivery Rate, Average Delivery Time, and any L3+ composite metric incorporating delivery time.\\[4pt]
\textbf{Exception:} Orders exceeding the 60-day threshold are still counted as \emph{late} for On-Time Delivery Rate---the exclusion applies only to the average used in internal benchmarking, not the binary late/on-time classification.\\[4pt]
\textbf{Expected Impact:} $\sim$0.3\% of orders affected; including these outliers skews Average Delivery Time upward by $\sim$1.5 days.
\end{tcolorbox}

\section{Ablation: Impact of Artifact Linkage Graph}
\label{app:ablation_ALG}

To isolate the graph's contribution to rule selection, we hold each task's skeleton fixed---the metric, dimensional slice, and time window---and vary only \emph{which rule is paired with the metric}. For every rule-impacting task ($N{=}175$), we re-pair the metric under three strategies and re-execute: \emph{Graph} uses the original graph-selected rule; \emph{Semantic} uses the nearest other rule by embedding cosine similarity; \emph{Random} uses a uniformly random other rule. A pairing is \emph{discriminative} if applying the rule changes the computed answer relative to the base (no-rule) query. Because the skeleton is identical across conditions, the comparison isolates the effect of rule selection alone.

\begin{table}[t]
\centering
\begin{tabular}{@{}lc@{}}
\toprule
\textbf{Pairing strategy} & \textbf{Discriminative validity} \\
\midrule
Graph    & 82\% \\
Semantic & 29\% \\
Random   & 22\% \\
\bottomrule
\end{tabular}
\caption{Discriminative validity under three rule-pairing strategies on identical task skeletons ($N{=}175$ rule tasks, both domains pooled). A pairing is discriminative if applying the rule changes the computed answer. The graph selects rules that genuinely modify the computation $\sim$3$\times$ more often than random pairing; semantic similarity helps only marginally, since a topically related rule need not alter the metric's value.}
\label{tab:kg_ablation}
\end{table}

The graph-selected rule changes the answer in 82\% of cases, versus 29\% for semantic-similarity pairing and 22\% for random pairing. The graph is not perfect---for some dimensional slices even a relevant rule leaves the answer unchanged (e.g., a delivery-outlier rule has no effect on a region with no outlier deliveries)---but it identifies discriminative pairings far more reliably than the alternatives. Semantic similarity offers only a small gain over random, confirming that topical relatedness does not ensure a rule \emph{causally modifies} the metric: the graph's typed rule$\rightarrow$metric edges, not embedding proximity, are what capture this relationship.

\section{Subgraph Scoring and Selection Details}
\label{app:sampling_details}

After enumerating structurally valid subgraphs (Section~\ref{sec:sampling}), the pipeline must select a diverse, high-quality subset. This involves two steps: (1) scoring each candidate, then (2) selecting via MMR to balance quality against diversity.

\subsection{Step 1: Quality Scoring}
\label{app:scoring}

Each candidate subgraph $S$ receives a composite quality score $q(S)$ as a weighted average of five independent scorers:
\begin{equation}
    \label{eq.scorers}
    q(S) = \sum_{i=1}^{5} w_i \cdot s_i(S)
\end{equation}
The five scorers capture complementary factors:

\begin{itemize}
    \item \textbf{Semantic affinity} (weight 0.25): Does the metric--dimension pairing make sense? For example, ``delivery time by customer state'' is natural, but ``delivery time by payment installment count'' is not. Although the graph constrains which dimensions are reachable from a metric's tables, not every reachable pairing is semantically meaningful; as such, this scorer downweights implausible ones.
    \item \textbf{Rule diversity} (weight 0.20): Prefers subgraphs using less-frequently-selected rule documents, via diminishing returns: $\frac{1}{|R_S|}\sum_{r \in R_S} \frac{1}{1+n_r}$, where $n_r$ counts prior selections of rule $r$ and $R_S$ is the set of rule documents.
    \item \textbf{Metric rarity} (weight 0.20): Analogous to rule diversity but for metrics---prefers subgraphs involving metrics that have appeared less often.
    
    \item \textbf{Dimension cardinality} (weight 0.20): How many distinct values does the chosen dimension have in the data? Dimensions with more values offer richer slicing possibilities, so the scorer increases monotonically: single-value dimensions score 0.1 (trivial questions), 2 values score 0.5 (only binary comparisons), 3--5 values score 0.8, and $>$5 values score 1.0. Date/timestamp columns (used for trend analysis) default to 0.7 because their effective cardinality depends on time-window granularity rather than the raw number of distinct dates.
    
    \item \textbf{Join complexity} (weight 0.15): Rewards subgraphs requiring more table joins, scaling linearly with the number of tables and capping at 1.0: one table scores 0.25, two tables 0.5, three tables 0.75, and four or more tables 1.0. This ensures the benchmark includes multi-join tasks.
\end{itemize}

Weights were determined by manual inspection: we adjusted until top-ranked candidates consistently matched subgraphs that a domain expert identified as realistic analysis questions.
\subsection{Step 2: MMR Selection}
\label{app:mmr}

High-quality candidates may still be redundant (e.g., many subgraphs testing the same metric with different dimension values). To ensure diversity, we use Maximal Marginal Relevance (MMR). Given candidate pool $\mathcal{C}$ and target count $k$, we greedily build the selected set $\mathcal{S}$:
\begin{equation}
    c^* = \arg\max_{c \in \mathcal{C} \setminus \mathcal{S}} \left[(1 - \lambda) \cdot q(c) + \lambda \cdot \min_{s \in \mathcal{S}} d_J(c, s)\right]
    \label{eq:mmr}
\end{equation}
where the first term favors quality and the second favors diversity. The distance $d_J$ is the Jaccard distance over node sets:
\begin{equation}
    d_J(a, b) = 1 - \frac{|V_a \cap V_b|}{|V_a \cup V_b|}
\end{equation}
with $V_a = \mathit{metrics}(a) \cup \mathit{tables}(a) \cup \mathit{dimensions}(a) \cup \mathit{rules}(a)$. Two subgraphs sharing no nodes have $d_J = 1$ (maximally diverse); identical subgraphs have $d_J = 0$.

We set $\lambda = 0.3$, favoring quality over diversity. This balances two failure modes: $\lambda$ too low produces redundant subgraphs testing the same metric--dimension pairs; $\lambda$ too high selects low quality subgraphs. For the first selection ($\mathcal{S} = \emptyset$), the diversity term defaults to 1.0, so the highest-quality candidate is chosen first. After MMR selection, a coverage pass adds any missing metrics or rule documents (those not yet represented in $\mathcal{S}$) by selecting their highest-scoring subgraph. 

In our pipeline, when setting the number of target questions to $100$, ${\sim}93\%$ of metric--rule links (averaged across datasets) are retained as parts of questions. The coverage loss is concentrated in metrics with many associated rules: once 2--3 subgraphs are selected for such a metric, the diversity term suppresses additional ones in favor of under-represented metrics.

\begin{table*}[ht]
\centering
\small
\begin{tabular}{@{}lp{2.2cm}p{2.8cm}p{2.2cm}p{3.2cm}c@{}}
\toprule
\textbf{Analysis Type} & \textbf{Metric(s)} & \textbf{Tables} & \textbf{dimension} & \textbf{Rules} & \textbf{Score} \\
\midrule
Point query & Late Delivery Rate by Carrier Segment & orders, customers, sellers, order\_items & customer\_state (spatial) & Active Seller Definition & 0.958 \\
\addlinespace
Trend analysis & Average Delivery Time & orders & order\_purchase
\_timestamp (temporal) & Outlier Exclusion (Delivery) & 0.807 \\
\addlinespace
Comparison & CLV, Revenue per Order, CSAT & customers, orders, order\_items & order\_status (categorical) & CLV Lifespan, CSAT Calculation, Revenue Recognition, \ldots (11 total) & 0.898 \\
\addlinespace
Ranking & Seller Fulfillment Score & order\_items, orders, order\_reviews & order\_status (categorical) & Seller Fulfillment Score Weights & 0.884 \\
\addlinespace
Significance testing & Payment Method Distribution & order\_payments & payment\_type (categorical) & Undefined Payment Exclusion & 0.879 \\
\addlinespace
Threshold crossing & Unique Buyer Count & orders, customers & order\_status (categorical) & Customer Deduplication & 0.846 \\
\addlinespace
Counterfactual & Category Growth Rate & order\_items, products, orders & order\_status (categorical) & CSAT Calculation, Multi-Item Attribution & 0.849 \\
\addlinespace
Decomposition & Seller Concentration, Active Seller Count, GMV & order\_reviews, orders, order\_items & order\_status (categorical) & Active Seller Definition, Revenue Recognition, \ldots (6 total) & 0.834 \\
\bottomrule
\end{tabular}
\caption{Representative selected subgraph for each analysis type from the Brazilian E-commerce domain. Scores reflect the composite quality metric $q(S)$ from Equation~\ref{eq.scorers}.}
\label{tab:subgraph_examples}
\end{table*}

\subsection{Subgraph Examples}
Table~\ref{tab:subgraph_examples} presents subgraph examples from the Brazilian E-commerce domain and Table~\ref{tab:subgraph_examples_cz} for the Czech Financial domain. 

\begin{table*}[ht]
\centering
\small
\begin{tabular}{@{}lp{2.2cm}p{2.8cm}p{2.2cm}p{3.2cm}c@{}}
\toprule
\textbf{Analysis Type} & \textbf{Metric(s)} & \textbf{Tables} & \textbf{dimension} & \textbf{Rules} & \textbf{Score} \\
\midrule
Point query & Transaction Volume & trans & operation (categorical) & Account Opening Date Rule & 0.866 \\
\addlinespace
Trend analysis & Balance Volatility (CV) & trans & trans.date (temporal) & Small Sample Suppression & 0.801 \\
\addlinespace
Comparison & Credit Risk Score & loan, trans, account, district, client & district region (spatial) & Credit Risk Score Weights & 1.000 \\
\addlinespace
Ranking & Average Account Balance & trans, account & k\_symbol (categorical) & Fee Structure Change 1996, Inter-Bank Transaction Flag & 0.863 \\
\addlinespace
Significance testing & Standing Order Value by Type & order & k\_symbol (categorical) & Standing Order Amount Validation & 0.858 \\
\addlinespace
Threshold crossing & Debt Service Coverage Ratio & trans, loan, account & k\_symbol (categorical) & Cohort Maturity Threshold, NPL Denominator & 0.849 \\
\addlinespace
Counterfactual & Card Penetration Rate & card, disposition, account & card type (categorical) & Card Eligibility Definition, Date Conversion & 0.889 \\
\addlinespace
Decomposition & Credit Risk Score, DSCR, Balance Volatility, NPL Ratio & district, loan, account, trans, client & district region (spatial) & Credit Risk Weights, Cohort Maturity, Crisis Adjustment 1997, \ldots (11 total) & 0.886 \\
\bottomrule
\end{tabular}
\caption{Representative selected subgraph for each analysis type from the Czech Financial domain.}
\label{tab:subgraph_examples_cz}
\end{table*}

\section{Validation Details}
\label{app:validation}

\subsection{Stage 1: Programmatic Validation}

The first stage applies deterministic checks. A task passes only if all checks return zero issues; failing tasks are flagged and annotated with the specific issues detected, but are excluded from LLM evaluation and the final benchmark.

\textbf{Structural integrity.} A set of regex patterns detects common generation artifacts: timestamp columns used as categorical filters (e.g., a question filtering on a datetime field as though it were a discrete value), raw column names leaked into the question text, and internal database identifiers such as primary or foreign key fields exposed in the phrasing. Questions shorter than 20 characters are also flagged. 

\textbf{Answer completeness.} Each task must have a valid answer. For SQL-based tasks, this requires a non-empty list of result dictionaries containing no error markers. For knowledge-only tasks, a substantive string answer of more than 10 characters must be present. For all non-knowledge tasks, the system additionally verifies that a scorable expected answer can be derived from the gold data, ensuring downstream automated scoring is feasible.

\subsection{Stage 2: LLM-Based Quality Evaluation}

Tasks (question-answer pairs) that pass programmatic validation are evaluated by an LLM judge (DeepSeek V3.2 via Amazon Bedrock) across eight dimensions with a binary score (0 or 1 each) and one-sentence justification per dimension. A task is admitted only if all dimensions pass. The prompt is as follows:

\begin{tcolorbox}[colback=gray!3, colframe=gray!50!black, title={\textbf{LLM Validator Prompt}}, fonttitle=\small, breakable]
\small\ttfamily
You are evaluating a question-answer-SQL triple from a benchmark dataset.\\[2pt]
ANALYSIS TYPE: \{analysis\_type\}\\
QUESTION: \{question\}\\
ANSWER: \{gold\_answer\}\\
SQL: \{gold\_sql or `NONE (knowledge-base task, no SQL needed)'\}\\
\{type\_specific\_context\}\\[4pt]
\rmfamily For each dimension, first write one sentence of reasoning, then give your score (1=pass, 0=fail). Only score 0 for CLEAR violations. When in doubt, pass.\\[4pt]
\begin{enumerate}\itemsep0pt
\item \textbf{OBJECTIVITY}: Does the question have exactly one correct answer given the data?
\item \textbf{ANSWER\_LEAKAGE}: Is the question free of hints that reveal the answer?
\item \textbf{NATURALNESS}: Does the question sound like a real person would ask it?
\item \textbf{ANSWER\_RELEVANCE}: Does the answer address the question's core intent? Structured tool outputs are valid.
\item \textbf{FORMAT\_ALIGNMENT}: Rankings must specify count. Breakdowns, significance, and trend always pass. Threshold: FAIL if question asks about multiple items (format holds one value only).
\item \textbf{SQL\_CORRECTNESS}: Does the SQL fetch appropriate data? Pass if SQL is NONE, if no date is mentioned and no date filter exists, or if tool-based SQL fetches raw data without performing the final computation.
\item \textbf{LOGICAL\_REASONING}: Is the answer internally consistent?
\item \textbf{GOLD\_SCOPE\_MATCH}: Does the answer contain ONLY data that the question asks about? FAIL if question asks about one entity but answer contains others. PASS if question asks for a breakdown and answer has multiple rows.
\end{enumerate}
\end{tcolorbox}

\subsection{Human-in-the-Loop Review}
\label{sec:hitl}

To assess benchmark quality, an expert annotator reviewed a stratified sample of 100 tasks across the same eight dimensions used by the automated validator. Table~\ref{tab:hitl} compares three views: the human ground-truth pass rate across all 100 tasks, the validator's pass rate on those same tasks, and the human-verified quality of the 25 tasks the validator admits (all-eight-pass).%

\begin{table}[t]
\centering
\setlength{\tabcolsep}{4pt}
\begin{tabular}{@{}lccc@{}}
\toprule
\textbf{Dimension} & \textbf{Human} & \textbf{Validator} & \textbf{Admitted} \\
                   & \textbf{(100)} & \textbf{(100)} & \textbf{(25)} \\
\midrule
Obj.        & 87\%  & 64\%  & 100\% \\
Leakage     & 100\% & 100\% & 100\% \\
Natural.    & 100\% & 100\% & 100\% \\
Relevance   & 92\%  & 83\%  & 100\% \\
Format      & 98\%  & 77\%  & 96\%  \\
SQL Corr.   & 94\%  & 48\%  & 100\% \\
Reasoning   & 100\% & 74\%  & 100\% \\
Scope       & 95\%  & 72\%  & 99\% \\
\midrule
Average     & 96\%  & 77\%  & \textbf{99\%} \\
\bottomrule
\end{tabular}
\caption{Human-in-the-loop validation (100-task stratified sample). \textbf{Human}: expert-annotator pass rate (ground truth); \textbf{Validator}: LLM validator pass rate on the same tasks; \textbf{Admitted}: human-verified pass rate of the 25 tasks the validator admits (all eight dimensions pass from the validator). The validator is stricter than the human across dimensions, so what it admits is near-perfect (99\% average). Its conservatism reduces yield (admits 25\% of candidates) but indicates high precision among validator-admitted tasks.}
\label{tab:hitl}
\end{table}

\subsection{Generator and Judge Sensitivity}
\label{app:generator_judge}
On a generator-swap sample with fixed skeletons and SQL-computed gold answers, Qwen3-80B- and DeepSeek-generated questions receive similar mean quality scores under an independent Claude Sonnet 4.6 judge (79.9\% versus 79.6\%), although individual dimensions move in both directions: SQL correctness decreases from 51\% to 34\%, while format alignment increases from 86\% to 99\%. The raw artifact contains 87 valid paired examples. Separately, four judges evaluated the same balanced human-labeled sample. Mean precision is 94--96\%, while recall ranges from 79--93\%; Opus has $n=98$ because two outputs could not be parsed. The judge is used for task admission, not for grading agent answers, which are scored against SQL-computed gold. These analyses bound but do not eliminate the shared-model confound.

\section{Comparison between Agents and Human Experts} 
\label{app:comparison_agent_human}
In order to interpret agent accuracy on DI-Bench, we need a human reference point. Without one, we cannot tell whether agents' low performance on certain tasks reflects an agent-human gap or a task that humans also find difficult.
We therefore sample 5 questions from each of the \emph{Knowledge}, \emph{Knowledge + Analytics}, and \emph{Knowledge + Analytics + Rule} tiers and ask two business-intelligence experts to answer them.
The experts were able to solve 13 out of 15 tasks (87\%) correctly, whereas the agents reach 73\% (Opus), 67\% (Haiku), 60\% (Qwen), and 53\% (DeepSeek) on the same samples. This shows that the benchmark differentiates agent capabilities and is not merely filled with tasks that are impossible to solve. The gap instead reflects tasks that humans can solve but agents struggle with.

\section{Benchmark Construction Time}
Our proposed pipeline substantially reduces the human effort required to construct a benchmark. We estimate that fully manual construction takes an expert roughly 35 minutes per task: subgraph creation (3 min), finding an example SQL query (4 min), writing and verifying the rule-modified SQL (15 min), phrasing the question (5 min), specifying the answer format (3 min), and a final check (5 min), or about 420 hours for a 731-task set.
With the pipeline, the only remaining human effort is verification, which, on average across all task types, takes roughly ~1.8 minutes or 22 hours for the entire benchmark, which is a significant time reduction from 420 hours required to construct the benchmark manually.

\section{Artifact Linkage Graph Statistics}
\label{app:kg_stats}

This appendix provides statistics on the number of nodes and edges of each defined type in the artifact linkage graphs, for both datasets. These are provided in Table~\ref{tab:alg_stats}.

\begin{table}[h]
\centering
\begin{tabular}{lcc}
\toprule
 & \textbf{Brazilian} & \textbf{Czech} \\
\midrule
Table nodes & 9 & 8 \\
Metric nodes & 29 & 29 \\
Rule nodes & 17 & 17 \\
Dimension nodes & 21 & 17 \\
Total nodes & 76 & 71 \\
\midrule
metric$\rightarrow$table edges & 61 & 45 \\
table$\rightarrow$dimension edges & 21 & 17 \\
table$\rightarrow$table edges & 9 & 8 \\
rule$\rightarrow$metric edges & 79 & 79 \\
metric$\rightarrow$metric edges & 9 & 11 \\
rule$\rightarrow$dimension edges & 12 & 8 \\
Total edges & 191 & 168 \\
\bottomrule
\end{tabular}
\caption{Artifact linkage graph statistics for both domains.}
\label{tab:alg_stats}
\end{table}

\section{Example Tasks}
\label{app:examples}

We present one representative task for each of the nine analysis types, drawn verbatim from the benchmark (questions, required rules, and gold answers unedited). The mark is Claude Opus 4.6's result.

\begin{tcolorbox}[colback=blue!3, colframe=blue!40!black, title={\textbf{Point Query}}, fonttitle=\small]
\small
\textbf{Question:} What is the average loan size and total number of loans for our 60-month active loans in the South Moravia region?\\[3pt]
\textbf{Required rules:} cohort maturity threshold; small-sample suppression\\[3pt]
\textbf{Gold:} \texttt{[\{region: south Moravia, duration: 60, avg\_loan\_size: 228422.0, loan\_count: 30\}]}\\[3pt]
\textbf{Opus:} \ding{51}
\end{tcolorbox}

\begin{tcolorbox}[colback=red!3, colframe=red!40!black, title={\textbf{Ranking}}, fonttitle=\small]
\small
\textbf{Question:} Which 5 customer states had the longest average delivery times in days for orders shipped during 2017 and the first eight months of 2018?\\[3pt]
\textbf{Required rules:} outlier exclusion (delivery); on-time delivery SLA\\[3pt]
\textbf{Gold:} \texttt{[AM: 25.65, AP: 24.76, AL: 24.12, RR: 23.76, PA: 23.0]} (avg delivery days)\\[3pt]
\textbf{Opus:} \ding{51}
\end{tcolorbox}

\begin{tcolorbox}[colback=purple!3, colframe=purple!40!black, title={\textbf{Comparison}}, fonttitle=\small]
\small
\textbf{Question:} Compare the Seller Concentration Index (HHI) for `computers\_accessories' against `cds\_dvds\_musicals' across all delivered and shipped orders. Which has higher concentration risk?\\[3pt]
\textbf{Required rules:} multi-item attribution; privacy/confidentiality\\[3pt]
\textbf{Gold:} \texttt{\{value\_a: 0.03, value\_b: 1.0, higher: cds\_dvds\_musicals\}}\\[3pt]
\textbf{Opus:} \ding{51}
\end{tcolorbox}

\begin{tcolorbox}[colback=green!3, colframe=green!40!black, title={\textbf{Trend Analysis}}, fonttitle=\small]
\small
\textbf{Question:} Show me the monthly trend for our average delivery time throughout 2017 --- any noticeable patterns or fluctuations?\\[3pt]
\textbf{Required rules:} outlier exclusion (delivery)\\[3pt]
\textbf{Gold:} \texttt{\{direction: increasing, total\_change\_pct: 32.1, peak\_period: 2018-01\}}\\[3pt]
\textbf{Opus:} \ding{51}
\end{tcolorbox}

\begin{tcolorbox}[colback=orange!3, colframe=orange!50!black, title={\textbf{Threshold Crossing}}, fonttitle=\small]
\small
\textbf{Question:} Did our average Seller Ship-Out Time in the state of Bahia exceed the 60-hour target?\\[3pt]
\textbf{Required rules:} S\~{a}o Paulo SLA change (January)\\[3pt]
\textbf{Gold:} \texttt{\{value: 58.33, threshold: 60.0, result: FAIL\}}\\[3pt]
\textbf{Opus:} \ding{51}
\end{tcolorbox}

\begin{tcolorbox}[colback=cyan!3, colframe=cyan!40!black, title={\textbf{Significance Testing}}, fonttitle=\small]
\small
\textbf{Question:} Are the average installment counts for orders paid by credit card vs.\ boleto in Q1 2018 statistically significantly different?\\[3pt]
\textbf{Required rules:} outlier exclusion (delivery)\\[3pt]
\textbf{Gold:} \texttt{\{significant: true, p\_value: 0.0, test: Welch's t-test (two-tailed)\}}\\[3pt]
\textbf{Opus:} \ding{51}
\end{tcolorbox}

\begin{tcolorbox}[colback=teal!3, colframe=teal!40!black, title={\textbf{Counterfactual}}, fonttitle=\small]
\small
\textbf{Question:} What's the average revenue per unique customer in S\~{a}o Paulo state, assuming we could extend operations there starting from January 2017?\\[3pt]
\textbf{Required rules:} none (control)\\[3pt]
\textbf{Gold:} \texttt{[\{customer\_state: SP, revenue\_per\_customer: 129.41\}]}\\[3pt]
\textbf{Opus:} \ding{51}
\end{tcolorbox}

\begin{tcolorbox}[colback=gray!8, colframe=gray!55!black, title={\textbf{Knowledge Retrieval}}, fonttitle=\small]
\small
\textbf{Question:} How do we calculate Total Deposit Volume from the transaction data?\\[3pt]
\textbf{Required rules:} transaction net-flow rule\\[3pt]
\textbf{Gold:} ``Total Deposit Volume must only sum amounts where \texttt{type = `PRIJEM'}.'' (must-include: \texttt{type = `PRIJEM'})\\[3pt]
\textbf{Opus:} \ding{51}
\end{tcolorbox}

\begin{tcolorbox}[colback=red!6, colframe=red!55!black, title={\textbf{Decomposition} (hardest type)}, fonttitle=\small]
\small
\textbf{Question:} Break down the Order Defect Rate by seller state for H1 2018, showing which geographic areas contribute most to the overall defect rate.\\[3pt]
\textbf{Required rules:} on-time delivery SLA; S\~{a}o Paulo SLA change (January)\\[3pt]
\textbf{Gold:} \texttt{[BA: 18.18, MS: 66.67, ES: 18.18, MA: 16.67, \ldots]} (18 states, odr\_pct)\\[3pt]
\textbf{Opus:} \ding{55}\ computed defect counts without the SLA grace period, so most rows are off.
\end{tcolorbox}

\section{Per-Model Accuracy by Analysis Type}
\label{app:per_model_type}

Table~\ref{tab:per_model_type} breaks down accuracy by model and analysis type over the full 731-task benchmark (BR and CZ combined). The difficulty ordering is consistent across models: all four exceed 75\% on knowledge retrieval and 60\% on threshold crossing and trend analysis, but fall below 35\% on ranking, point query, counterfactual, and decomposition. Model separation is largest on the harder types, e.g.\ on point query, Opus (34\%) clearly leads Qwen3-80B (18\%) and Haiku (22\%)---confirming that the discriminative signal concentrates in tasks requiring exact multi-row matching and multi-rule reasoning.

\begin{table*}[t]
\centering
\begin{tabular}{lrcccc}
\toprule
\textbf{Analysis Type} & \textbf{N} & \textbf{Opus} & \textbf{Haiku} & \textbf{Qwen} & \textbf{DeepSeek} \\
\midrule
Ranking             & 145 & 23 & 20 & 11 & 23 \\
Knowledge retrieval & 137 & 89 & 85 & 78 & 76 \\
Threshold crossing  & 123 & 72 & 61 & 65 & 67 \\
Point query         & 102 & 34 & 22 & 18 & 23 \\
Trend analysis      &  78 & 78 & 63 & 78 & 78 \\
Comparison          &  63 & 79 & 59 & 65 & 65 \\
Sig.\ testing       &  42 & 71.4 & 64.3 & 57.1 & 59.5 \\
Counterfactual      &  21 &  9.5 &  4.8 &  9.5 &  4.8 \\
Decomposition       &  20 &  5.0 &  5.0 &  5.0 &  5.0 \\
\midrule
\textbf{Overall}    & 731 & \textbf{57.9} & 49.0 & 47.9 & 50.9 \\
\bottomrule
\end{tabular}
\caption{Per-model accuracy (\%) by analysis type on the full 731-task benchmark (Brazilian e-commerce and Czech financial combined). N is the number of tasks of each type.}
\label{tab:per_model_type}
\end{table*}
\section{Detailed Failure Cases}
\label{app:failure_cases}

To make concrete \emph{what} a task expects and \emph{why} it is hard, we walk through three Claude Opus 4.6 failures from the Brazilian e-commerce domain. All three are tasks where the agent successfully \emph{retrieved} the governing rule, as verified in the tool trace, yet still produced a wrong answer isolating rule \emph{application} as the bottleneck rather than retrieval. Expected values are the deterministic SQL-computed gold; predicted values are taken verbatim from the agent's \texttt{<answer>} tag.

\begin{tcolorbox}[colback=orange!3, colframe=orange!50!black, title={\textbf{Case 1 --- Threshold crossing flips on a one-day grace period (\texttt{DI-BR-0117})}}, fonttitle=\small, breakable]
\small
\textbf{Question:} For delivered orders in the first half of 2018, did the order defect rate for sellers in the Federal District (DF) exceed 15.0\%?\\[3pt]
\textbf{Required rules:} on-time delivery SLA; delivery outlier exclusion\\[3pt]
\textbf{Expected:} \texttt{\{value: 14.46, threshold: 15.0, result: FAIL\}}\\[3pt]
\textbf{Opus:} \ding{55}\ \texttt{\{value: 16.40, threshold: 15.0, result: PASS\}}\\[3pt]
\textbf{Why it is hard:} The SLA rule classifies an order as late only if it arrives \emph{after the end of the next business day} following the estimated date---a one-business-day grace period. Opus retrieved the rule but counted any delivery past the estimated date as a defect (\texttt{delivered\_date > estimated\_date} rather than \texttt{> estimated\_date + 1 day}). The grace period removes just enough borderline-late orders to drop DF's ODR from 16.40\% to 14.46\%---across the 15\% threshold. Because the metric is compared against a threshold, a sub-two-point numeric error flips the categorical verdict from FAIL to PASS; the task is unforgiving of an approximately-right computation.
\end{tcolorbox}

\begin{tcolorbox}[colback=red!3, colframe=red!40!black, title={\textbf{Case 2 --- Ranking requires a post-hoc privacy suppression (\texttt{DI-BR-0123})}}, fonttitle=\small, breakable]
\small
\textbf{Question:} Which product categories have the highest concentration of sales among sellers? Provide a ranking of the top 5 categories by the Seller Concentration Index (HHI), considering only those where market dominance is most extreme.\\[3pt]
\textbf{Required rules:} multi-item attribution; data confidentiality\\[3pt]
\textbf{Expected:} \texttt{[la\_cuisine: 1.0, home\_comfort\_2: 1.0, fashion\_sport: 1.0, cds\_dvds\_musicals: 1.0, arts\_and\_craftmanship: 1.0]}\\[3pt]
\textbf{Opus:} \ding{55}\ \texttt{[cds\_dvds\_musicals: 1.0, la\_cuisine: 0.88, computers: 0.58, security\_and\_services: 0.54, flowers: 0.48]}\\[3pt]
\textbf{Why it is hard:} An HHI of 1.0 means a category's sales come from a single seller. The confidentiality rule requires suppressing data points derived from fewer than three unique sellers \emph{after} the base metric is computed, and the multi-item attribution rule fixes how each order's revenue is assigned to categories. Opus computed HHI correctly but never applied the suppression-and-tie logic, returning one HHI=1.0 category followed by lower-concentration ones, missing that several categories tie at the maximum 1.0 and that the ranking among ties is rule-governed. The rule operates as a post-processing layer applied \emph{after} SQL aggregation, which the agent treated as irrelevant once it had numeric HHI values.
\end{tcolorbox}

\begin{tcolorbox}[colback=teal!3, colframe=teal!40!black, title={\textbf{Case 3 --- Counterfactual demands two parallel computations (\texttt{DI-BR-0314})}}, fonttitle=\small, breakable]
\small
\textbf{Question:} What would our On-Time Delivery Rate look like if we removed the grace period for all orders delivered in the first half of 2018, broken down by customer state?\\[3pt]
\textbf{Required rules:} delivery outlier exclusion; week/timezone definition\\[3pt]
\textbf{Expected (excerpt):} \texttt{AL: \{strict: 72.73, with\_grace: 74.68\}}, \texttt{AM: \{strict: 94.55, \ldots\}}\\[3pt]
\textbf{Opus (excerpt):} \ding{55}\ \texttt{BA: \{strict: 81.29, with\_grace: 82.85\}}, \texttt{CE: \{strict: 74.13, with\_grace: 76.06\}}\\[3pt]
\textbf{Why it is hard:} The agent must compute the same metric twice: once with the grace period (the live rule) and once without (the counterfactual scenario), while applying the outlier-exclusion filter consistently to both. A single question thus requires two parallel, rule-aware computations over the same rows, and any inconsistency in how either rule is applied causes the per-state values to diverge from gold.
\end{tcolorbox}

\paragraph{Takeaway.} All three failures share a structure: retrieval succeeds, the base SQL is roughly right, but the rule-induced adjustment (a grace period, a suppression rule, a counterfactual rule swap) is either dropped or misapplied. This matches the aggregate finding (Section~\ref{sec:experiments}) that the with-rule analytical tasks are the benchmark's primary discriminator.

\section{Pipeline Reproduction}
\label{app:reproduction}
DI-Bench takes a relational schema and tables, a metric catalog with executable SQL templates, and a policy-document corpus. Figure~\ref{fig:pipeline_reproduction} documents the implementation details: graph construction, candidate-scoring weights, MMR selection, the base/rule-aware SQL record, validation inputs, rare-type augmentation, and stage commands. Appendix~\ref{app:validation} gives the complete validator prompt.
\begin{figure*}[t]
\centering
\tcbset{
  pipelinebox/.style={
    listing only,
    listing engine=listings,
    colback=gray!7,
    colframe=gray!35,
    boxrule=0.4pt,
    arc=1pt,
    left=4pt,
    right=4pt,
    top=3pt,
    bottom=3pt,
    fonttitle=\bfseries\scriptsize,
    colbacktitle=gray!15,
    coltitle=black,
    listing options={
      basicstyle=\ttfamily\scriptsize,
      breaklines=true,
      breakatwhitespace=false,
      columns=fullflexible,
      keepspaces=true,
      showstringspaces=false
    }
  }
}
\begin{tcbraster}[
  raster columns=2,
  raster equal height=rows,
  raster column skip=2mm,
  raster row skip=2mm
]
\begin{tcblisting}{pipelinebox,title={A. Graph construction}}
graph = Graph(nodes={table, metric, rule, dimension})
graph.add(schema_edges(schema))
# table-table and table-dimension
graph.add(metric_edges(catalog))
# metric-table and metric-metric
graph.add(infer_rule_links(documents, catalog))
# rule-metric and rule-dimension
pool = connected_subgraphs(graph)
pool = [g for g in pool if structural_constraints_hold(g)]
\end{tcblisting}
\begin{tcblisting}{pipelinebox,title={B. Candidate scoring and selection}}
weights = {
  semantic: .25,
  rule_diversity: .20,
  metric_rarity: .20,
  cardinality: .20,
  joins: .15
}
quality = score_candidates(pool, weights)
chosen = greedy_mmr(pool, quality, target=100,
                    diversity=.30, seed=42)
chosen = ensure_metric_and_rule_coverage(chosen, pool)
\end{tcblisting}
\begin{tcblisting}{pipelinebox,title={C. SQL grounding and question generation}}
for g in chosen:
  for kind in compatible_analyses(g):
    base_sql = instantiate_template(g.metric,
                                    g.dimensions, kind)
    rule_ops = extract_rule_operations(
      g.rules,
      allowed={filter, exclude, threshold,
               weight, override})
    gold_sql = apply_operations(base_sql, rule_ops)
    base = enrich(execute(base_sql), kind)
    gold = enrich(execute(gold_sql), kind)
    if failed(gold) or (rule_ops and equivalent(base, gold)):
      continue
    question = generate_question(g, kind, gold, documents)
    task = record(question, kind, g.id, base_sql,
                  gold_sql, rule_ops, gold, g.rules)
\end{tcblisting}
\begin{tcblisting}{pipelinebox,title={D. Validation, augmentation, and export}}
    if not programmatic_checks(task):
      continue
    judge_input = {question, gold, gold_sql,
                   type_context(kind)}
    dimensions = {objectivity, leakage, naturalness,
                  relevance, format, correctness,
                  reasoning, scope_match}
    # Full prompt appears in the Validation Details appendix
    if all(binary_judge(judge_input, d)
           for d in dimensions):
      tasks.append(task)

tasks = augment_to_minimum(tasks,
          unused=pool-chosen, minimum=10, seed=42)
tasks = deduplicate(tasks, keys={task_id, subgraph_id})
export(tasks)
\end{tcblisting}
\end{tcbraster}
\vspace{2mm}
\begin{tcblisting}{pipelinebox,title={E. Stage commands}}
$ python kg_construction/extract_graph.py
$ python kg_construction/extract_policy_edges_llm.py all --model deepseek.v3.2
$ python subgraph_sampling/sample_subgraphs.py --all --target 100 --seed 42
$ python task_generation/generate_tasks.py --all --seed 42 --use-llm
$ python task_generation/enrich_tasks.py --all
$ python task_generation/generate_questions.py --all --use-llm
$ python task_generation/validate_questions.py --all --use-llm
$ python task_generation/augment_rare_types.py --all --target 10 --use-llm
\end{tcblisting}
\caption{Five-panel reference implementation: graph construction, weighted candidate selection, SQL grounding and backward generation, validation and export, and executable stage commands. Policy-bearing tasks are retained only when rule removal changes the executed answer.}
\label{fig:pipeline_reproduction}
\end{figure*}
\paragraph{Reproduction boundary.} The listing is sufficient to reproduce the generation procedure, but not necessarily the exact task wording: LLM-backed stages depend on model availability and nondeterministic decoding. The released configuration records model identifiers and decoding parameters, while non-LLM randomized stages use seed 42.

\end{document}